\documentclass[prodmode,acmtsas]{acmsmall}
\usepackage{amssymb}
\usepackage{amsmath}
\usepackage{graphicx}
\usepackage{epstopdf}
\usepackage{algorithm}
\usepackage{algpseudocode}
\usepackage{pifont}
\usepackage{url}
\usepackage{tabularx}
\usepackage{makecell}
\usepackage{multirow}
\usepackage{tabularx}
\usepackage{ragged2e}
\usepackage{booktabs}
\usepackage{caption}
\usepackage[table]{xcolor}

\newcommand{\In}{\textbf{in }}

\newcommand{\Is}{\textbf{is }}
\graphicspath {{figures/}}

\begin{document}
\markboth{H. Chen et al.}{Geo-referencing Places from Everyday Natural Language Place Descriptions}
\title{Geo-referencing Places from Everyday Natural Language Place Descriptions}
\author{Hao Chen
\affil{University of Melbourne}
Maria Vasardani
\affil{University of Melbourne}
Stephan Winter
\affil{University of Melbourne}}

\begin{abstract}
Natural language place descriptions in everyday communication provide a rich source of spatial knowledge about places. An important step to utilize such knowledge in information systems is geo-referencing all the places referred to in these descriptions. Current techniques for geo-referencing places from text documents are using place name recognition and disambiguation; however, place descriptions often contain place references that are not known by gazetteers, or that are expressed in other, more flexible ways. Hence, the approach for geo-referencing presented in this paper starts from a place graph that contains the place references as well as spatial relationships extracted from place descriptions. Spatial relationships are used to constrain the locations of places and allow the later best-matching process for geo-referencing. The novel geo-referencing process results in higher precision and recall compared to state-of-art toponym resolution approaches on several tested place description datasets. 
\end{abstract}



\acmformat{Hao Chen, Maria Vasardani, and Stephan Winter, 2016. Geo-referencing Places from Everyday Natural Language Place Descriptions.}
\begin{bottomstuff}
Author's addresses: H. Chen, M. Vasardani, and S. Winter, Department of Infrastructure Engineering, University of Melbourne, Parkville, Victoria 3010, Australia.
\end{bottomstuff}
\maketitle

\section{Introduction}
With the increasing volume of unstructured text documents being published online, as well as the growing need for place-related information in everyday life, the relation between places and text documents has recently attracted research attention, and the necessity of identifying and locating places from text documents has been emphasized by others \cite{jones2001geographical,schlieder2001qualitative,hill2006georeferencing,teitler2008newsstand}. Place information extracted from text can be used to facilitate a wide range of applications such as geographic information retrieval \cite{silva2006adding,purves2007design,jones2008geographical}, to smooth human-computer interaction \cite{raubal2009cognitive,winter2016place,davies2009user}, and to build place information systems. The rapid development of text mining and natural language processing techniques makes it feasible to extract information from text documents, through information extraction techniques such as named entity recognition and relation extraction.

This research focuses on natural language place descriptions as data input. Natural language place descriptions occur in everyday verbal communication as a way of encoding and transmitting spatial knowledge about place between individuals \cite{vasardani2013locating,vasardani2013descriptions} as well as in written texts such as web documents, news articles, social media texts, trip guides, and tourism articles \cite{teitler2008newsstand,kim2015harvesting}. Such place descriptions provide a qualitative reference system for describing geographic locations, and consist essentially of references to places and the qualitative spatial relations between these places. Consider for example the following transcription of an emergency call:

\begin{quotation}
	``We need an ambulance. We are in the Cussonia Courtyard, on the campus. The courtyard is beside the clock-tower. The closest road is probably Monash Road. You can rush through the Old Quad.''
\end{quotation}

The information conveyed by such a place description is useful for mental sketching of a spatial environment, and can be used, for example, to provide navigational instructions or to inform the location of events. 

Current techniques for geo-referencing place from text documents, i.e., linking place references to geographic locations or footprints, are based on toponym resolution \cite{leidner2007toponym} which relies on the identification of place names in external knowledge bases (typically a gazetteer) and performing disambiguation if there is a reference ambiguity \cite{delozier15:gazetteer,lieberman12:adaptive,garbin*05:disambiguating,gouvea*08:discovering,li06exploring,smart10multi,buscaldi2008conceptual,buscaldi2008map}. However, places extracted from everyday place descriptions are challenging for this approach. First, everyday place descriptions are flexible in language, and often contain place references that cannot be found in a gazetteer. For example, these references can be synonyms to gazetteered names, such as vernaculars (e.g., `FedSquare' instead of `Federation Square'), references of otherwise limited spread (e.g., `the place where we met yesterday'), or categories (e.g., `the train station'), possibly with additional qualifiers (e.g., `the central station'). Many of the vernacular descriptions refer to places of vague boundaries (e.g., `the BBQ area on the lawn'). Such places in place descriptions can be located only by the provided spatial relationships to other places. Secondly, places in everyday place descriptions are frequently of a spatial granularity where environmental features are no longer gazetteered. Individual buildings, establishments in buildings (e.g. rooms), or features of local interest (e.g., ATMs) usually have higher ambiguity than larger geographic features thus require different approaches to disambiguate and resolve \cite{buscaldi2011approaches}. Current toponym resolution approaches, designed for larger geographic features such as populated places (e.g., cities or countries) or natural geographic features (e.g., rivers or mountains) can, for example, use heuristics based on the sizes of the features (e.g., population). Such an approach is not applicable for everyday features that are too numerous and too similar. In summary, not as much attention and effort has been spent on developing approaches for resolving place references that are fine-grained or cannot be found in a gazetteer -- which are common in everyday place descriptions. 

This research aims at overcoming these limitations in order to geo-reference \emph{all} places in everyday place descriptions. The paper presents a methodology that starts from a graph containing extracted places and their spatial relationships: a \emph{place graph} \cite{vasardani2013descriptions}. The hypothesis of this research is that integrating the structure of the place graph into the geo-referencing process allows to exceed the state-of-art toponym resolution approaches in precision and recall. The approach presented in this paper has been implemented and tested successfully. Hence, the contributions of this paper can be summarized as follows:
\begin{enumerate}
	\item We propose a new toponym resolution approach based on a place graph with merged information extracted from any number of (everyday) place descriptions;
	\item We demonstrate how all places from a place graph can be geo-referenced even if some are expressed by place references that cannot be recognized by a gazetteer;
	\item We evaluate our approach with experiments on different datasets based on precision and recall and compare the result to state-of-art toponym resolution approaches.
\end{enumerate}
The remainder of the paper is structured as follows: In Section 2 a review of related work is given. Section 3 clarifies how place references and spatial relationships are modelled by a place graph, as well as the roles they play in the following geo-referencing approach. In Section 4, a multi-step geo-referencing approach is explained. Section 5 shows implementation and experiment results on several test datasets. In Section 6 a discussion is presented. Section 7 concludes.

\section{Related work}
People talk about space by referring to places \cite{winter2010spatially}. Bennett \emph{et al}. define places as conceptual entities that enable cognitive structuring of the spatial aspects of reality \cite{bennett2007semantic}. In GIScience and related geographic research fields, the notion of place is a central concept in human spatial cognition and communication \cite{tuan1977space,goodchild2011formalizing}. Place based research is an emerging research dimension in GIScience in order to smooth and simplify human-computer interaction through capturing, modeling and utilizing place-related information, and the importance of place based research has been widely acknowledged (e.g., \cite{golledge1997spatial,goodchild2007citizens,goodchild11:formalizing,winter2012approaching,winter2016place}). 

Identifying and locating place from unstructured text documents has recently attracted place based research attention. In the remaining parts of this section, relevant tools for geo-referencing place from text documents will be introduced.

\subsection{Gazetteer}
In order to locate place names on a map with precise coordinates, gazetteers are often used in conjunction to maps. A gazetteer is an important component in geo-referencing systems for both enterprise and academic purpose, and is commonly used for geographic information retrieval, navigation services and web-mapping applications. A gazetteer typically contains three core components: place names, feature types, and footprints \cite{hill2000core,goodchild2008introduction}, and is often regarded as a geospatial dictionary of geographic names. A place name is what people usually search for this place, and is typically considered as `the official name'. A place type is a category from a feature-type thesaurus for classifying places according to their semantics. A footprint represents the location of a place, typically by a single coordinate pair as an estimated center of an extended object, and sometimes by a polygon or a polyline instead. Some gazetteers, such as the Getty Thesaurus of Geographic Names (TGN)\footnote{http://www.getty.edu/research/tools/vocabularies/tgn/} or GeoNames\footnote{http://www.geonames.org/}, also store alternative names, and provide detailed descriptions of places as well as positions of places in administrative or political hierarchies.

\subsection{Toponym resolution}
The goal of toponym resolution is to recognize place names from text documents and link them to geographic locations or footprints, and the essential challenges are place name recognition and disambiguation \cite{leidner2007toponym}. Disambiguation is the process of mapping each place name to its actual geographic locations when there is more than one candidate reference locations. For example, according to GeoNames, the toponym `Paris' can refer to more than sixty different geographic locations around the world.

Disambiguation approaches can be classified into map-based (e.g. \cite{smith2001disambiguating,zhang2012disambiguating}), knowledge-based (e.g. \cite{buscaldi2008conceptual,karimzadeh2013geotxt}), and machine learning (e.g., \cite{smith2003bootstrapping,garbin2005disambiguating}). Various heuristics have also been suggested based on features such as population and whether a place name is a capital city name \cite{leidner2007toponym}. The selection of the disambiguation approach is usually based on the task and data source available \cite{buscaldi2011approaches}. Geotagging systems -- typically systems that determine the geo-focus for the entire document for geographic information retrieval purposes -- use toponym resolution techniques existing in literature or customized ones (e.g., \cite{teitler2008newsstand,lieberman2007steward}).

Existing toponym resolution approaches are not suitable for the task of this research due to three reasons. First, these approaches typically focus on gazetteered place names, while everyday places descriptions often contain place references that are not gazetteered. Second, place descriptions may contain vague places that can only be geo-referenced using spatial relationships to other places. Third, these existing approaches typically focus on places of spatial granularities that are larger or equal to suburb- and city-level, which are easier to resolve than places of finer spatial granularities. Fine-grained places often require additional information and a different methodology to disambiguate and resolve. 

\subsection{Qualitative spatial relationships}
Qualitative spatial relationships reflect spatial cognitive capacity of people \cite{vasardani2013locating} and provide useful knowledge for understanding locative expressions. Qualitative spatial relationships have been extensively studied in the Artificial Intelligence community for qualitative spatial reasoning, including cardinal \cite{freksa1992using,frank1992qualitative,liu2005internal}, topological \cite{egenhofer1991point,randell1992spatial}, relative direction \cite{schlieder1995reasoning,freksa1992using}, and qualitative distance \cite{frank1992qualitative,worboys2001nearness}. Spatial relationships can be modeled using formal logic and applied for tasks such as robotic navigation. In English, such qualitative spatial relationships are often expressed by prepositions thus can be identified and extracted from text documents.

Some studies use qualitative spatial relationships to derive uncertainty fields for locative expressions \cite{herskovits1985semantics}. People use locative expressions to describe a vague location through spatial relationships to some known place. E.g., `10$km$ east of Berkeley' refers to some unspecified place at a 10$km$ distance in a particular direction from a known place called Berkeley. Wieczorek \emph{et al.} developed such a point-radius method using cardinal directions and (imprecise) metric distance relationships \cite{wieczorek2004point}. Associated uncertainties such as coordinate-, distance-, and direction-imprecision are calculated in order to derive an uncertainty field representation. The methodology was later modified by applying a probabilistic distribution model, as the possibilities of a place to be located at any location within the uncertainty field are not equal \cite{guo2008georeferencing}. Liu \emph{et al.} go a step further by adding topological and qualitative distance relationships in the model \cite{liu2009positioning}. 

Other studies attempt to quantify qualitative spatial relationships. A study by Delboni \emph{et al.} focus on determining semantic equivalence for spatial relationships through quantification \cite{delboni2007semantic}. Fu \emph{et al.} assigns different search radii for \emph{near} based on the semantic categories of the referred places given a spatial query \cite{fu2005ontology}. Hall \emph{et al.} conduct a series of data-driven studies to quantify spatial relationships including \emph{near} and cardinal direction relationships, in terms of distance and orientation \cite{hall2008quantifying,hall2008evaluating}. The approach by Skoumas \emph{et al.} \cite{skoumas2016location} is comparable to the one proposed by Hall \emph{et al.}, which uses grid-based representations for the derived probabilistic models for spatial relationships.

The above approaches focus on either deriving uncertainty probabilistic fields for spatial relationships, or on the quantification of spatial relationships. Yet how qualitative spatial relationships can be used in the task of geo-referencing, i.e., linking a place referred by spatial relationships to specific geographic locations or the footprint stored by a gazetteer, remains undiscussed. In addition, these previous studies typically assume some already geo-referenced relata, and in our case relata are not geo-referenced beforehand.

\subsection{Place graph} \label{pg}
A locative expression can be modelled by a \emph{locatum}, the reference to a place that is to be located, a \emph{relatum}, the reference to a place that is already located, and a spatial relationship between the two. Such representation, e.g., \textless building, near, south lawn\textgreater, has been called a spatial triplet \cite{vasardani2013descriptions}. Approaches for extracting triplets from place descriptions are available \cite{khan2013extracting,liu2014automatic}.

A spatial property graph $G=(V, E)$ can then be constructed based on a set of triplets, and a complete construction process is given by Kim \emph{et al.} \cite{kim2015descriptions}. Each triplet is stored as two nodes, one each for locatum and relatum, and an edge in between for the spatial relationship. Spatial relationships are standardized, e.g., `to the north of' and `Northern' are both normalized by an edge `north of'. Each edge in the graph are directed from locatum to relatum due to the asymmetry of spatial relationships, and there can be multiple edges for different spatial relationships between the same pair of nodes. Nodes are merged through a comprehensive similarity matching process considering string, linguistic and spatial relationship \cite{kim2016similarity}. If multiple place references are identified to be referring to the same place by their similarities, they are stored as a single node, i.e., each node has a unique identifier and potentially multiple place references. 

All places from such a place graph are not yet geo-referenced. The approach in this research starts from a place graph as data input, instead of raw place descriptions as text documents which are typically used by toponym resolution studies.

\section{Place graph as a knowledge base}
Before explaining the geo-referencing approach, this section clarifies how places, place references and spatial relationships are modeled by a place graph.

\subsection{Place and place reference} \label{reference}
Places are referred to in place descriptions by place references. People use a variety of ways to refer to places; examples were given above. Between places and place references are many-to-many relationships, i.e., a place may be referred to by one or more different place references, while the same place reference may be used to describe different places in different context (e.g., `the train station in Melbourne central').

In this research, places and place references are further categorized by whether they are gazetteered or not. Regarding \emph{places}, a \textit{gazetteered place} is a place that is stored by a gazetteer, and a \textit{non-gazetteered} place is a place that is not stored by a gazetteer. Regarding \emph{place references}, a \textit{gazetteered reference} is a place name stored in a gazetteer as the name of a gazetteered place, and a \textit{non-gazetteered reference} is a place reference that is not known by a gazetteer. Examples for illustration will be given later.

Naturally, a place reference to a gazetteered place can be different to its gazetteered name. For example, two references `Flinders Street Railway Station' (gazetteered) and `the train station' (non-gazetteered) come from conversational contexts where they referred to the same, gazetteered place (Flinders Street Railway Station). In a different context, the reference `the train station' may refer to another train station. 

\subsection{Place in a place graph} \label{Place}
\begin{figure}[ht]
	\centering
	\includegraphics[width=\textwidth]{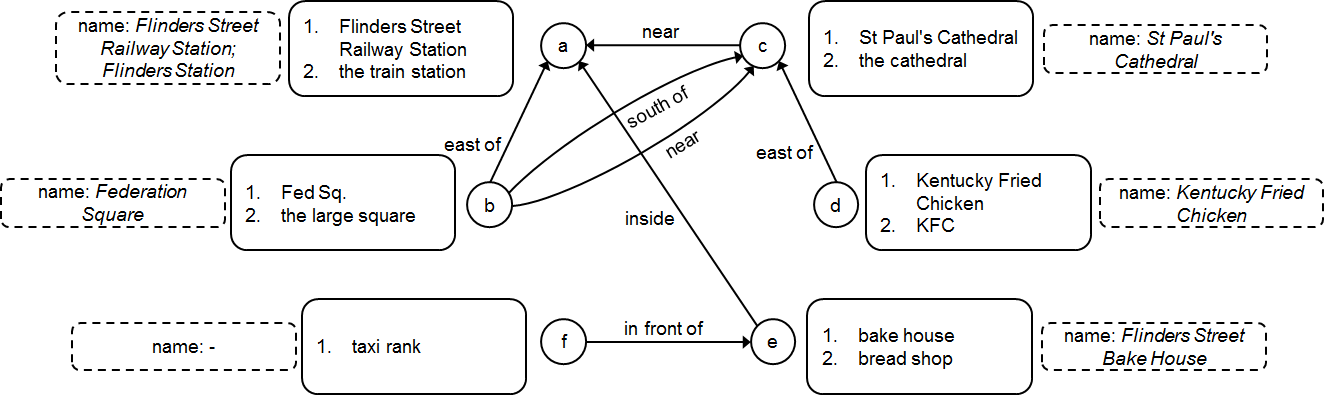}
	\caption{A sample place graph with six nodes and seven edges}
	\label{fig:place graph}
\end{figure}

Figure~\ref{fig:place graph} shows a sample place graph constructed from everyday place descriptions. It consists of six places represented by nodes (labeled \textit{a, b, c, d, e, f}) as well as seven spatial relationships represented by labeled edges. A list of place references that have been found in various place descriptions for each node is shown in the solid-line rectangles. Each dashed-line rectangle shows the ground-truth gazetteered name(s) for these places (`-' for non-gazetteered places). The ground-truth names are only shown for demonstration purposes, and are not available from the input place graph.

Such a place graph may be derived from multiple place descriptions, e.g., a place graph of Melbourne constructed from hundreds of place descriptions about Melbourne as Web documents. The construction methodology is given in Section~\ref{pg}. The geo-referencing approach below starts from such a place graph, while current toponym resolution processes typically focus on a single document at a time.

In the sample place graph shown in Figure~\ref{fig:place graph}, nodes \textit{a, b, c, d, e} are gazetteered places, and node \textit{f} is a non-gazetteered place. Each of those nodes may have several place references, and each of them may be a gazetteered reference. Thus, three situations can be distinguished for places in a place graph, which lead to a multi-step geo-referencing approach.
\begin{enumerate}
	\item A gazetteered place has at least one gazetteered reference (nodes \textit{a, c, d});
	\item A gazetteered place has no gazetteered references (nodes \textit{b, e}); and
	\item A non-gazetteered place (node \textit{f}).
\end{enumerate}

\subsection{Qualitative spatial relationships as constraints} 
Spatial relationships provide valuable knowledge for inferring the approximate location of a place by connecting it to other places, e.g., `the bake house inside Flinders Street Station'. The inferred approximate location can then be used for gazetteer matching and resolving reference ambiguity. In this research, the semantics of qualitative spatial relationships from four families are considered, as shown in Table~\ref{table:rel}. In order to be able to use these families studied in Artificial Intelligence, a mapping of the much richer, more flexible, and deeply contextualized natural language of spatial prepositions and verbs indicating relationships has to be applied to place graphs.

\begin{table}[ht]
	\centering
	\caption{Spatial relationships considered in the approach below}
	\label{table:rel}
	\begin{tabularx}{\textwidth}{l X}
		\Xhline{4\arrayrulewidth}
		Spatial relationship family 						& Spatial relationships \\
		\hline
		Cardinal direction				& north of, south of, east of, west of, north east of, south east of, north west of, south west of \\
		
		Qualitative distance			& near \\
		
		Relative direction				& in front of, behind, left of, right of \\
		
		Topological					& inside, covered by, overlap, meet, disjoint, cover, contain, equal \\
		\Xhline{4\arrayrulewidth}
	\end{tabularx}
\end{table}

A \textit{search space} is defined for each relationship from the four families to represent the constrained location of a locatum that satisfies the spatial relationship to an already geo-referenced relatum. In the geo-referencing approach below, search spaces are used for filtering out gazetteer entries that do not satisfy those given spatial relationships. Thus, for each locatum, only a constrained number of gazetteer entries will be left for the later best-matching process.

\subsubsection{Cardinal direction relationships} 
\begin{figure}[ht]
	\centering
	\includegraphics[width=0.7\textwidth]{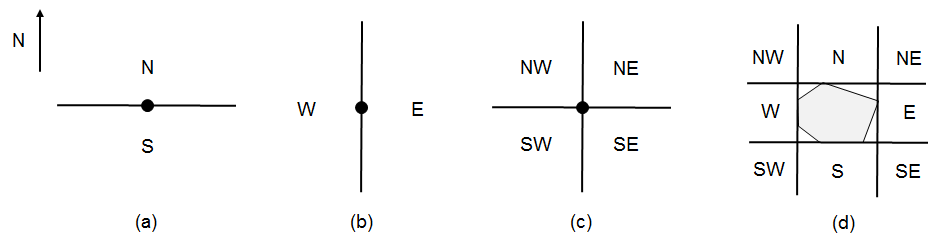}
	\caption{Search spaces for cardinal direction relationships}
	\label{fig:cd}
\end{figure}

The search spaces for cardinal direction relationships are defined in this paper using Frank's half-plane models \cite{frank1992qualitative} if the relatum is geo-referenced as a point, as shown in Figure~\ref{fig:cd} (a), (b) and (c). If the relatum is geo-referenced as a polygon, its centroid will be used to derive half-planes. An alternative model for polygon-based relata is minimal-bounding-box based (Figure~\ref{fig:cd} (d)). This paper applies the former (centroid-based) model since a cardinal direction relationship from a place description may be an internal cardinal relationship, e.g., `in the north (northern part) of the city', while using the latter model will result in misinterpretation in such cases.

\subsubsection{Topological relationships} 
\begin{figure}[ht]
	\centering
	\includegraphics[width=0.5\textwidth]{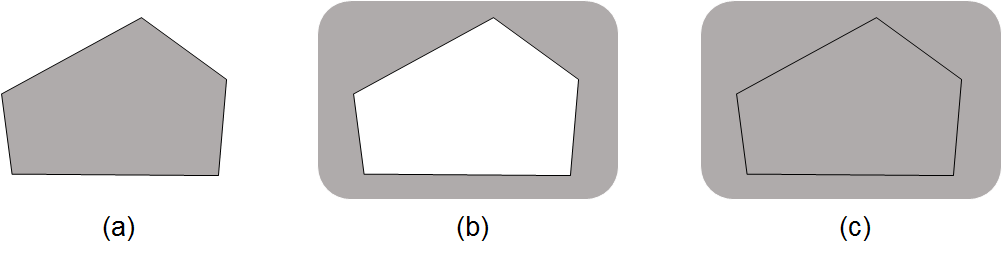}
	\caption{Search spaces for \textit{covered by, equal, inside} (a), \textit{disjoint, meet} (b), and others (c)}
	\label{fig:to}
\end{figure}

If the relatum is geo-referenced as a polygon, the search spaces for topological relationships are defined as shown in Figure~\ref{fig:to}. If the relatum is geo-referenced as a point or polyline, no search spaces will be defined. As shown by the figure, only the search spaces for \textit{covered by, equal} and \emph{inside} are limited, while the search spaces for other topological relationships can not be constrained by limited areas directly. In natural language people often use containment relationships for expressing topological relationships (typically \emph{inside}), therefore in most cases the derived search spaces for topological relationships are constrained. 

For all relationships other than \textit{covered by, equal} and \emph{inside}, the defined search spaces are not limited. However, once all gazetteer entries that are within a search space are obtained, they can further be filtered through computing their geometries (for polygons only) and the geometry of the relata to determine whether these entries satisfy the given topological relationship. The detailed process for filtering by geometrical computation will be explained in Section~\ref{spatial}. 

\subsubsection{Qualitative distance relationship} \label{qualitative}
\begin{figure}[ht]
	\centering
	\includegraphics[width=0.7\textwidth]{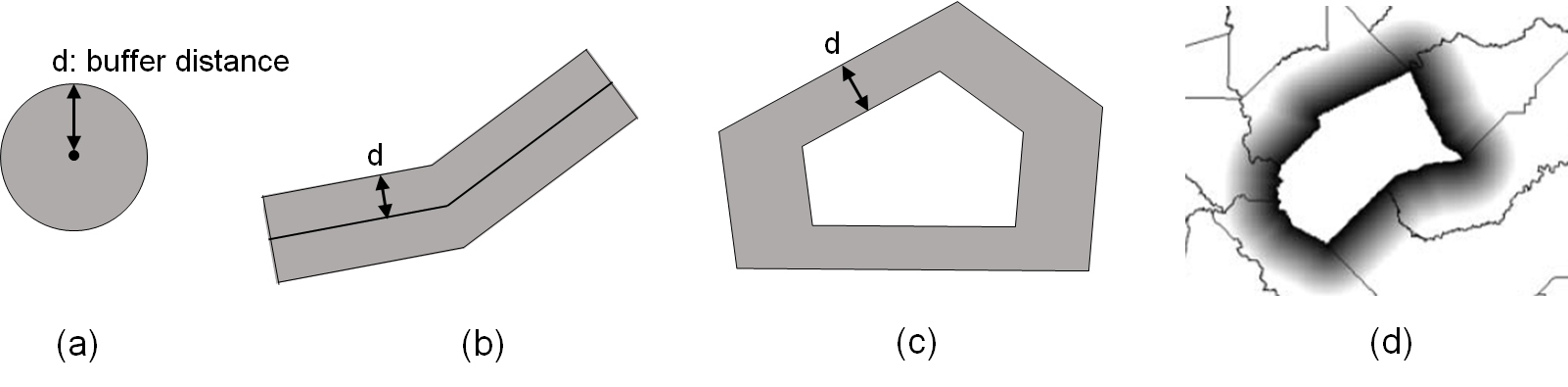}
	\caption{Search space for \emph{near} in this research (a), (b), (c) and by Liu \emph{et al}. (d)}
	\label{fig:qd}
\end{figure}

The search space for the qualitative distance relationship \emph{near} is defined in a comparable way to the probabilistic uncertainty model proposed by Liu \emph{et al.} \cite{liu2009positioning}, and a comparison is shown in Figure~\ref{fig:qd}. The search space in this research is a buffered region based on the geometry of the relatum, and can be computed using existing buffering algorithms. The computation of nearness in order to deal with the probabilistic uncertainty of \emph{near} will be given in Section~\ref{spatial}. 

Defining a buffer distance is the general accepted model for quantifying the qualitative nearness in local-search applications as well as geographic information retrieval engines (e.g., \cite{fu2005ontology}). In this research, the buffer distance is defined considering the size of involved objects (relata) as well as the scale of the spatial context, as shown in Eq.~\ref{eq:near}. $d$ stands for the buffer distance, $\alpha$ is a constant, and $\beta, \gamma$ are two coefficients that makes $d$ positively correlated with the area of the relatum as well as the area of the spatial context. The computation of spatial contexts will be explained in Section~\ref{dis}. Different parameters values will be tuned in the implementation stage, and associated results will be discussed. 

\begin{equation}
\label{eq:near}
	d = \alpha + \beta \ast area(relatum) + \gamma \ast area(spatialContext)
\end{equation}

Another qualitative spatial relationship \emph{far} is not considered since it offers little help in constraining the location of a given locatum, as the search space would be unlimited. Furthermore, unlike some relationships which also have unlimited search spaces, e.g., \emph{overlap}, \emph{far} can not be used for filtering out candidate gazetteer entries.

\subsubsection{Relative direction relationships} 
\begin{figure}[ht]
	\centering
	\includegraphics[width=0.43\textwidth]{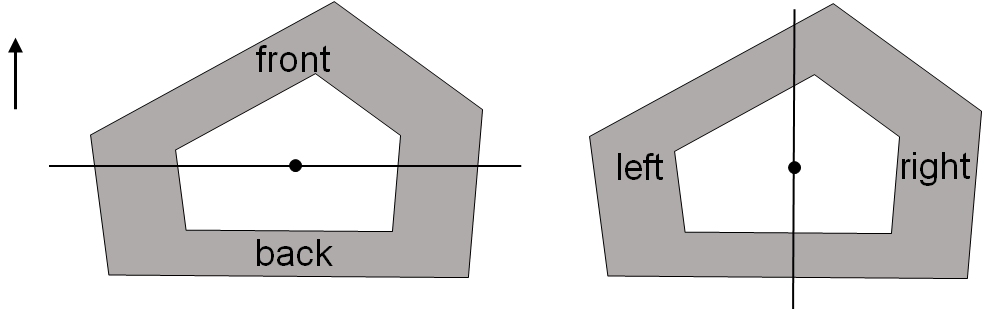}
	\caption{Search spaces for relative directions given a reference frame}
	\label{fig:rd}
\end{figure}

Search spaces for relative direction relationships are defined in a way similar to both cardinal direction and qualitative distance relationships. Relative direction relationships are based on orientation reference frames used by people, and can be either deictic, intrinsic or extrinsic \cite{retz1988various}. Assuming that the reference frame used is known, the search spaces could be defined as shown in Figure~\ref{fig:rd}. The arrow in the figure shows the direction of `front' used by the reference frame. Search spaces are defined by the intersection of half planes centered at the centroid of the relatum, and the search space of \textit{near}, as relative direction relationships are mostly used for describing places that are near to each other. 

No existing natural language parser is able to infer the reference frame automatically from place descriptions unless explicitly given by the description provider. If information of the reference frame used is unavailable, the search spaces for relative direction relationships are defined the same as for \textit{near}, as a fall-back approach.

\subsubsection{Approximate location region} 
\begin{figure}[ht]
	\centering
	\includegraphics[scale=0.25]{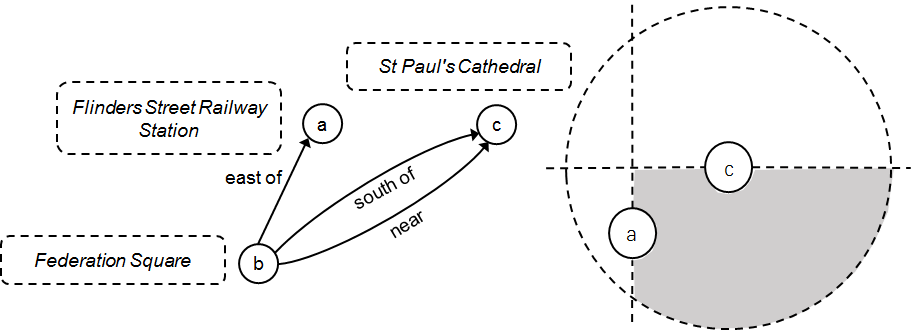}
	\caption{An example of deriving ALR for place \textit{b} through intersection of search spaces}
	\label{fig:ALR}
\end{figure}

An \emph{approximate location region} (ALR) is a derived region that represents the approximate location of a place based on all known spatial relationships to some already geo-referenced places, and is computed by the intersection of all search spaces for this place. For instance, as shown in Figure~\ref{fig:ALR}, place \textit{b} from the sample graph has no gazetteered references; however three outgoing relationships are available, i.e., east of \textit{a}, south of and near \textit{c}. Assuming that \textit{a} and \textit{c} are already geo-referenced, the location of \textit{b} can be constrained by the shaded region representing the ALR.

In the geo-referencing approach below, ALRs are used to limit the likely location of places as well as to limit the number of candidate gazetteer entries to be obtained for the later best-matching process. As a place graph gets populated with more place descriptions and spatial relationships, the ALRs generated are expected to become closer to the actual footprints thus increase the overall geo-referencing accuracy.

\section{Approach for geo-referencing}
The three situations described in Section~\ref{Place} for all places in a place graph are addressed by a three-step approach accordingly. The first step is to identify, disambiguate and geo-reference places in the place graph that have at least one gazetteered reference (Situation 1). These places are then regarded as \textit{anchor places}. For each of the places that do not have associated gazetteered references (Situation 2), a comprehensive best-matching process is conducted based on all stored spatial relationships to the previously geo-referenced anchor-places. Gazetteer entries with the highest overall similarity scores will be selected for geo-referencing. Finally, places that cannot be matched during the second step, i.e., non-gazetteered place (Situation 3), will be geo-referenced using their derived ALRs.

The input for the following algorithms is a place graph, a labelled directed multi-graph represented by a set of $n$ places $\{p_{1},p_{2},\ldots,p_{n}\}$ and a set of $k$ relationships $\{r_{1},r_{2},\ldots,r_{k}\}$ between these places. Each place $p_{i}$ corresponds to a node in the graph and is associated with a number of place references $\{p_{i}^1,p_{i}^2,\ldots,p_{i}^m\}$, $m \ge 1$. Each relationship is a tuple consisting of the starting node, the spatial relationship, and the ending node, e.g., $r_{1}=(p_{2}, east\_of, p_{1})$.

\subsection{Geo-referencing anchor places} \label{anchor}
\subsubsection{Identification of anchor places}
The algorithm for identifying anchor places is shown in Algorithm~\ref{FG}. The function \textit{getReferences} takes a place (e.g., $p_{i}$) as input and returns a list of all associated references of this place (i.e., $\{p_{i}^1,p_{i}^2,\ldots,p_{i}^m\}$). \emph{getGazetteerEntries} retrieves all gazetteer entries that exactly match a place reference in string, and returns a list (empty if the place references are non-gazetteered).

\begin{algorithm}[ht]
	\caption{Identifying anchor places and obtaining gazetteer entries}
	\label{FG} 
	\textbf{Input:} \emph{PlaceList} $\{p_{1},p_{2},\ldots,p_{n}\}$ \\
	\textbf{Output:} \emph{AnchorPlacesAndEntries}
	\begin{algorithmic}
		\State \emph{AnchorPlacesAndEntries} $:= \emptyset$
		\For {$p_{i}$ \In \emph{PlaceList}}								\Comment{for each place}
		\State \emph{Entries} $:= \emptyset$
		\For {$p_{i}^m$ \In getReferences($p_{i}$)}	\Comment{for each place reference of this place}
		\State \emph{Entries} $\longleftarrow$ getGazetteerEntries($p_{i}^m$)
		\EndFor
		\If {\emph{Entries} $\neq \emptyset$}			\Comment{if the place has at least one gazetteered reference}	
		\State \emph{AnchorPlacesAndEntries} $\longleftarrow$ ($p_{i}$, \emph{Entries})
		\EndIf
		\EndFor
		\State \Return \emph{AnchorPlacesAndEntries}
	\end{algorithmic}
\end{algorithm}

Essentially, Algorithm~\ref{FG} looks up all place references in an input place graph using a gazetteer. If a place has at least one associated place reference that can be found in the gazetteer, it is regarded as an anchor place. 

Taking the sample graph as an example, each of node \textit{a, c, d} has at least one gazetteered reference thus will be identified as an anchor place by the algorithm. The next step is disambiguation, as some place references may have more than one entries in the gazetteer, e.g., `St Paul's Cathedral, Melbourne; St Paul's Cathedral, London; St Paul's Cathedral, Bendigo;...'.

\subsubsection{Disambiguation} \label{dis}
Knowledge-based disambiguation approaches are not suitable for this research since place descriptions often contain fine-grained places, and the coverage provided by the commonly used external knowledge resources, e.g., Wikipedia and WordNet, are very limited for such places. Also, data-driven disambiguation approaches can not be applied since it is difficult to obtain annotated training corpus for such fine-grained places. In addition, some commonly used heuristics such as those based on population or jurisdiction level, are not applicable either.

\begin{figure}[ht]
	\centering
	\includegraphics[scale=0.18]{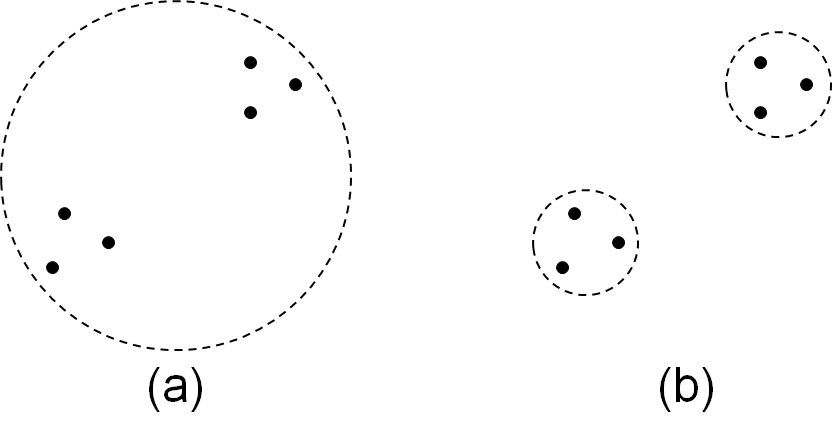}
	\caption{Comparison of clustering by (a) overall minimal distance, and (b) point density}
	\label{fig:cluster}
\end{figure}

Thus, a novel disambiguation approach is proposed: a density-based clustering approach that is superior to the previous (map-based) approaches at least for the task at hand. For example, some studies (e.g., \cite{habib2012improving,buscaldi2008map}) use overall-minimal-distance for disambiguation. If the data source contains places from multiple spatial foci that are away from each other, e.g., places from two suburbs or cities, their clustering algorithms will generate one large cluster. In contrast, the algorithm proposed in this research will generate multiple small clusters covering the foci separately. A comparison is illustrated in Figure~\ref{fig:cluster}. Note that the dashed circular regions are not indicating the actual cluster boundaries. Stronger disambiguation, or smaller clusters, are more useful for limiting the locations for other places in the following steps. Moncla \emph{et al.} \cite{moncla2014geocoding} use an existing clustering algorithm called DBSCAN \cite{ester1996density} for disambiguation. However, the algorithm requires manual input of the parameters of the neighborhood radius, $\varepsilon$, and the minimum number of points required to form a dense region, \emph{MinPts}, which in the case of Moncla \emph{et al.} are empirically adjusted based on the dataset. In contrast, the proposed algorithm is parameter-free. Requiring manually defined parameters makes DBSCAN unrobust for place graphs, as place graphs can be of various spatial scopes and extents. 

The clustering algorithm in this research is inspired by Ripley's $K$ function \cite{ripley1976second} which was originally designed to assess the degree of departure of a point set from complete spatial randomness, ranging from spatial homogeneity to a clustered pattern. Spatial randomness is irrelevant in this research, yet detecting point density meets our interest. A \textit{distance interval} $K$ function is developed, as shown in Eq.~\ref{eq:dk}. It computes the overall point density within the region of a given distance interval $(d-\Delta d, d]$ for all points. Here $no.[p\in region(p_{i}, {(d-\Delta d, d]})]$ represents the number of points that are at a distance between $d-\Delta d$ and $d$ from point $p_{i}$. The denominator of the left side of the function is the area of the region, and the left side part is used for computing point density. $\Delta d$ is used for discretizing the function, and by default is set to 100$m$. The original $K$ function can be regarded as a cumulative version of the distance interval $K$ function.

\begin{equation}
\label{eq:dk} 
K(d) = \frac{1}{\pi d^{2} - \pi (d-\Delta d)^{2}} \times \frac{1}{n}\sum_{i=1}^{n}no.[p\in region(p_{i}, {(d-\Delta d, d]})]
\end{equation}

\begin{figure}[ht]
	\centering
	\includegraphics[scale=0.32]{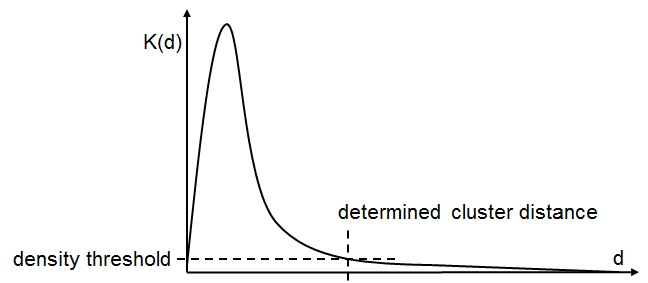}
	\caption{Deriving the cluster distance based on a distance interval $K$ function}
	\label{fig:k}
\end{figure}

The goal of the function is to detect clusters in the input point cloud that have significantly larger point densities, as well as to derive a cluster distance for deriving clusters. As shown in Figure~\ref{fig:k}, $K(d)$ increases sharply at the beginning, indicating at least one cluster. To determine the density threshold, the average density value $\overline{K(d)}$ and the standard deviation $\sigma$ of $K(d)$ for all discrete distance intervals are calculated. Then the 3$\sigma$ rule is adopted and the density threshold is $\overline{K(d)} + 3\sigma$. The complete disambiguation process using distance interval $K$ function is shown in Algorithm~\ref{cdk}.

\begin{algorithm}[ht]
	\caption{Disambiguation using distance interval $K$ function}
	\label{cdk} 
	\textbf{Input:} \emph{AnchorPlacesWithEntries} $\{(p_{i}, entries_{i}),\ldots\}$ \\
	\textbf{Output:} \emph{DisambiguatedAnchorPlaces} $\{(p_{i}, entry_{i}),\ldots\}$
	\begin{algorithmic}
		\State \emph{PointCloud} $:= \emptyset$  \Comment{all obtained entries as input point cloud}
		\For {\emph{Entry} \In \emph{AnchorPlacesWithEntries}.getAllEntries()}
			\State \emph{PointCloud} $\longleftarrow$ \emph{Entry}.getCoordinates()
		\EndFor
		
		\\ 													\Comment{calculate distance interval $K$ function}
		
		\State \emph{MaxDistance} $:=$ maximumPointWiseDistance(\emph{AllEntries})
			\For {$d$ \In iterate(0, \emph{MaxDistance}, $\Delta d$)} \Comment{loop of (min, max, interval)}
				\State $K(d) :=$ clusterDistanceKFunction(\emph{AllEntries})
			\EndFor
		\State \emph{Threshold} $:= $ average($K(d)$) + $3 \times$ standardDeviation($K(d)$)
		
		\\												\Comment{determine cluster distance}

		\State \emph{ArgmaxDistance} = getArgmaxDistance($K(d)$)
		\State \emph{SatisfyingDistances} := $\emptyset$
		\For {$d$ \In range(0, \emph{MaxDistance}, $\Delta d$)}
			\If {$K(d) \geq$ \emph{Threshold} \textbf{and} $d \geq$ \emph{ArgmaxDistance}}							
				\State \emph{SatisfyingDistances} $\longleftarrow d$
			\EndIf			
		\EndFor
		\State \emph{ClusterDistance} $:=$ min(\emph{SatisfyingDistances})
		
		\\														\Comment{derive all clusters based on the cluster distance and disambiguation}
		
		\State \emph{DisambiguatedAnchorPlaces} $:= \emptyset$
		\State \emph{RankedClusters} $:=$ rank(computeClusters(\emph{ClusterDistance}))			
		\For {\emph{Cluster} \In \emph{RankedClusters}}					
			\For {\emph{AnchorPlace} \In \emph{AnchorPlacesWithEntries}}	\Comment{try disambiguation}
				\For {\emph{Entry} \In \emph{AnchorPlace}.getEntries()}
					\If {\emph{Entry} \In \emph{Cluster}}							
						\State \emph{DisambiguatedAnchorPlaces} $\longleftarrow$ (\emph{AnchorPlace}, \emph{Entry})
						\State \emph{AnchorPlacesWithEntries}.remove(\emph{AnchorPlace})
					\EndIf													
				\EndFor
			\EndFor	
		\EndFor		
		
		\State \Return \emph{DisambiguatedAnchorPlaces}
	\end{algorithmic}
\end{algorithm}

The first part of Algorithm~\ref{cdk} collects coordinates of the centroid of all retrieved gazetteer entries for all anchor places as the input for the distance interval $K$ function. In the second and third parts of the algorithm, the input point clouds are analyzed using the distance interval $K$ function to derive the cluster distance. \emph{getArgmaxDistance} returns $d$ where $K(d)$ is maximum. In the last part, all points within the cluster distance from each other are classified into one cluster (by function \emph{computeClusters}). Clusters with a single point will be discarded, and no overlapping of clusters is possible. All clusters are ranked based on the number of points contained (in decreasing order). Finally, all anchor places are disambiguated through iterating the ranked clusters. If an anchor place, for example, has no entry in the top ranking cluster, the second-ranking cluster will be tested, and continue until an entry is found.

\begin{figure}[ht]
	\centering
	\includegraphics[scale=0.25]{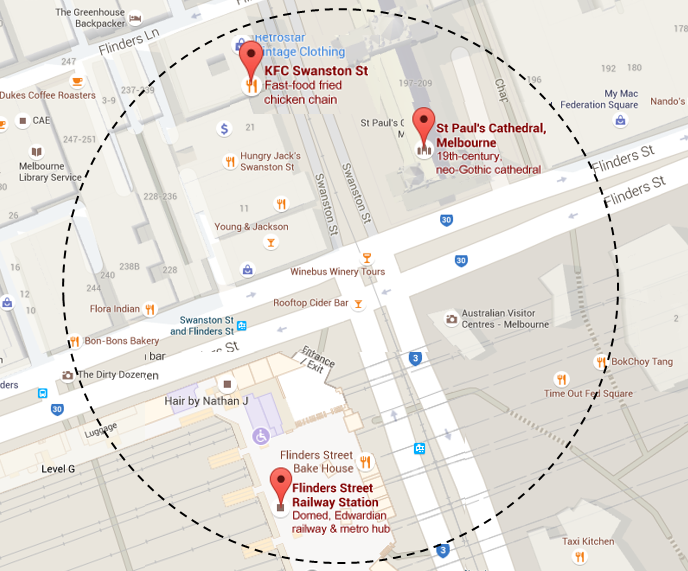}
	\caption{Example of disambiguated anchor places (Google Maps, 2016)}
	\label{fig:cdk}
\end{figure}

The disambiguation result for the sample input place graph is shown in Figure~\ref{fig:cdk}. Note that the dashed circular region is for illustration only and does not indicate the actual cluster boundary. Each cluster represents an approximate spatial scope where the original descriptions are geographically embedded, and the minimal bounding box of all the points within the cluster is regarded as the spatial context. The spatial context is used to decide the search space of \emph{near} for all anchor places within it, as mentioned in Section~\ref{qualitative}. The area of a spatial context can be different sizes, e.g., as several street blocks, suburb-, city- or state-level.

The robustness of the clustering algorithm will be tested through case studies. The computational intensive part of Algorithm~\ref{cdk} is calculating $K(d)$ for each distance interval, including calculating the point-wise distance for all entries as coordinates and binary search of the sorted distance list. 

\subsubsection{Further disambiguation}
It is possible that a derived cluster contains more than one entry for an anchor place. For example, assume the cluster region shown in Figure~\ref{fig:cdk} contains two entries of `Kentucky Fried Chicken'. In such a situation further disambiguation is needed. Remaining ambiguous anchor places will be temporarily excluded from anchor place list, and will be geo-referenced together with the remaining places in the next step through deriving ALRs. An ambiguous anchor place can be detected if it is associated with multiple entries in the \emph{DisambiguatedAnchorPlaces} list returned by Algorithm~\ref{cdk}.

\subsection{Geo-referencing places without gazetteered references through best-matching} \label{sec:bm}
The process of geo-referencing in this step is illustrated in Algorithm~\ref{BM}. The first part of the algorithm derives an ALR for each of the remaining places based on all the stored spatial relationships with the place as the starting node and an anchor places as the ending node. Candidate gazetteer entries, i.e., all gazetteer entries within the ALR regardless of name match, are then obtained. In the second part of the algorithm, a best-matching process is conducted based on reference similarity as well as spatial similarity. At the end of the matching process, each place will be geo-referenced with the candidate entry with the highest overall similarity score.

\begin{algorithm}[ht]
	\caption{Best-matching}
	\label{BM} 
	\textbf{Input:} \emph{DisambiguatedAnchorPlaces} $\{(p_{i}, entry_{i}),\ldots\}$, \\ \emph{RemainingPlaces} $\{p_{j},\ldots\}$, \emph{SpatialRelationships} $\{r_{1},r_{2},\ldots,r_{k}\}$ \\
	\textbf{Output:} \emph{BestMatchedPlaces} $\{(p_{j}, entry_{j}),\ldots\}$
	\begin{algorithmic}
		\State \emph{BestMatchedPlaces} $:= \emptyset$
		\For {$p_{i}$ \In \emph{RemainingPlaces}}		\Comment{derive an ALR and obtain gazetteer entries}
		\State \emph{Relationships} $:=$ getRelationshipsToAnchorPlaces($p_{i}$)
		\State \emph{ALR} $:=$ deriveALR(\emph{Relationships})
		\State \emph{Entries} $:=$ getGazetteerEntriesWithinALR(\emph{ALR}) \Comment{get entries by region}
		
		\\					\Comment{best matching}
		
		\State \emph{HighestSim} $:= 0$
		\For {\emph{Entry} \In \emph{Entries}}
		\State \emph{SpatialSim} $:=$ calculateSpatialSimilarity($p_{i}$, \emph{Entry})
		\For {$p_{i}^m$ \In getReferences($p_{i}$)}				\Comment{consider all stored place references}
		\State \emph{ReferenceSim} $:=$ calculateReferenceSimilarity($p_{i}^m$, \emph{Entry})
		\State \emph{OverallSim} $:=$ calculateOverallSimilarity(\emph{ReferenceSim}, \emph{SpatialSim})
		\If {\emph{OverallSim} $>$ \emph{HighestSim}}
		\State \emph{HighestSim} $:=$ \emph{OverallSim}
		\State \emph{BestEntry} := \emph{Entry}
		\EndIf	
		\EndFor
		\EndFor
		\State \emph{BestMatchedPlaces} $\longleftarrow$ ($p_{i}$, \emph{BestEntry}, \emph{HighestSim})
		\EndFor
		\State \Return \emph{BestMatchedPlaces}
	\end{algorithmic}
\end{algorithm}

\begin{figure}[ht]
	\centering
	\includegraphics[scale=0.24]{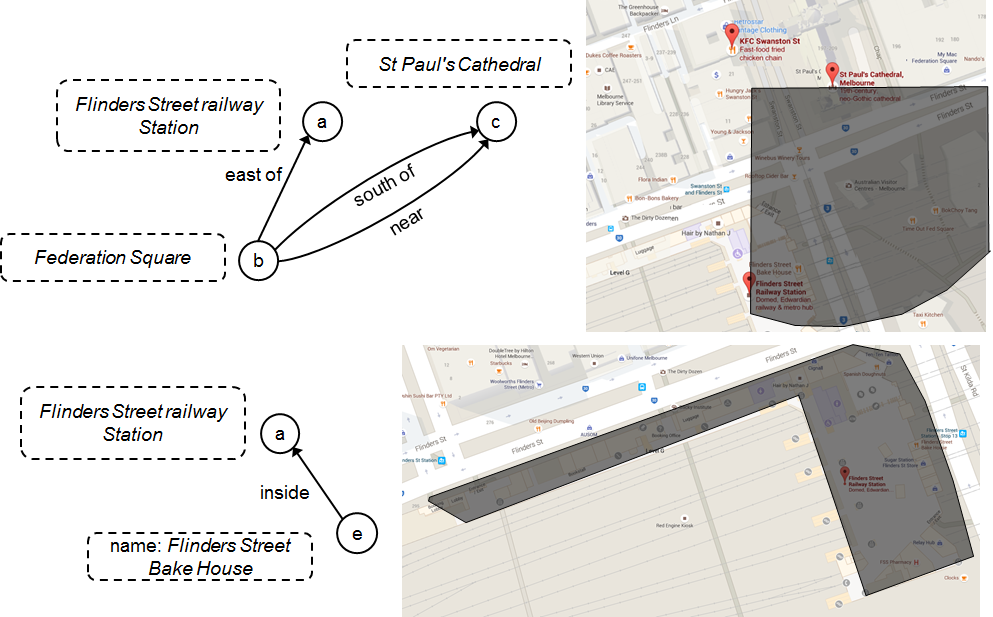}
	\caption{Derived ALRs for node \textit{b, e} (Google Maps, 2016)}
	\label{fig:alrsample}
\end{figure}

Taking the sample graph as an example, node \textit{b} and \textit{e} have some spatial relationships to anchor places \textit{a} and \textit{c}, and the ALRs derived for node \textit{b} and \textit{e} are illustrated in Figure~\ref{fig:alrsample}. All gazetteer entries within each of the derived ALR (shaded region shown in the figure) are obtained as candidate entries to be matched with. Note that node \emph{a} is assumed to be geo-referenced by a polygon to allow a search space for \emph{inside}.

\subsubsection{Reference similarity}
Reference similarity measures how well a candidate entry matches a place reference, both stringwise and semantically. String similarity measurement, e.g., based on edit-distance, does not consider word semantics, such as abbreviations (e.g., `bldg.' and `building') and words with similar meanings (e.g., `woods' and `forest', `department' and `section'). Furthermore, word re-ordering should also be considered (e.g., `St Paul's Cathedral' and `Cathedral of St Paul's'). Some gazetteers (e.g., OpenStreetMap\footnote{https://www.openstreetmap.org/}) also store tagging information associated with each gazetteer entry, e.g., \{name: \emph{Richard Berry}; type: \emph{building}; organization: \emph{University of Melbourne}; faculty: \emph{science}; department: \emph{Mathematics and Statistics}\}. Such tagging information is also useful for entry-matching if available. For instance, the Richard Berry building in the University of Melbourne is often referred as `school of mathematics and statistics' or simply `mathematics building' instead of its official name. Thus, the proposed method for reference similarity measurement is shown in Algorithm~\ref{sim}.

\begin{algorithm}[ht]
	\caption{function calculateReferenceSimilarity()}
	\label{sim} 
	\textbf{Input:} \emph{PlaceReference}, \emph{CandidateEntry} \Comment{both as string} \\
	\emph{SemanticSimilarityDictionary} 	\Comment{abbreviations and word-wise semantic similarity}
	\textbf{Output:} \emph{ReferenceSimilarity}
	\begin{algorithmic}
		\State \emph{PlaceReferenceTokens} $:=$ tokenize(\emph{PlaceReference})
		\State \emph{CandidateEntryTokens} $:=$ tokenize(\emph{CandidateEntry})
		\State \emph{TagRecordTokens} $:=$ tokenize(\emph{CandidateEntry}.getTagValues()) \Comment{if available}
		
		\\
		\State \emph{TokenSimList} $:= \emptyset$
		\For {\emph{PToken} \In \emph{PlaceReferenceTokens}}		\Comment{Find maximum match for each token}
			\State \emph{HighestSim} $:= 0$
			\For {\emph{CToken} \In \emph{CandidateEntryTokens}}		\Comment{compare reference and entry}
				\State \emph{TokenSim} $:=$ \emph{SemanticSimilarityDictionary}.getSim(\emph{PToken}, \emph{CToken})
				\If {\emph{TokenSim} $>$ \emph{HighestSim}}
					\State \emph{HighestSim} $:=$ \emph{TokenSim}
				\EndIf
			\EndFor
			\For {\emph{TToken} \In \emph{TagRecordTokens}}				\Comment{compare reference and tag if available}
				\State \emph{TokenSim} $:=$ \emph{SemanticSimilarityDictionary}.getSim(\emph{PToken}, \emph{TToken})
				\If {\emph{TokenSim} $>$ \emph{HighestSim}}
					\State \emph{HighestSim} $:=$ \emph{TokenSim}
				\EndIf
			\EndFor
			\State \emph{TokenSimList} $\longleftarrow$ \emph{HighestSim}
		\EndFor 
		\State \emph{ReferenceSimilarity} $:=$ average(\emph{TokenSimList})	\Comment{average similarity of token pairs}
		\State \Return \emph{ReferenceSimilarity}
	\end{algorithmic}
\end{algorithm}

The function \emph{SemanticSimilarityDictionary.getSim($token\_1$, $token\_2$)} returns the semantic similarity between two tokens, e.g., `department' and `building', and the function can be implemented using WordNet synsets as lexicons \cite{miller1995wordnet}. Some algorithms and implementations already exists (e.g., \cite{ballatore2013grounding}). Common abbreviations are considered as having $1.0$ similarity to the original words. If the similarity value cannot be found in the dictionary for two input tokens, a fall-back measurement is based on Damerau-Levenshtein edit distance similarity \cite{damerau1964technique}.

\begin{figure}[ht]
	\centering
	\includegraphics[scale=0.35]{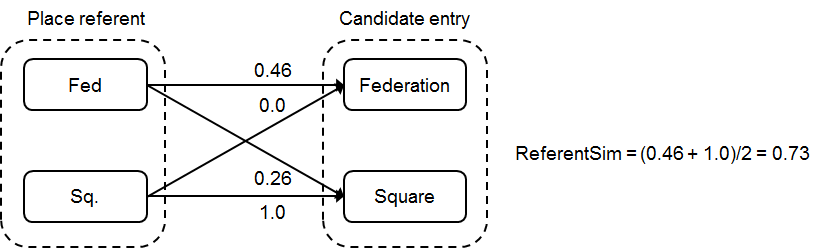}
	\caption{Computing the reference similarity between `Fed Sq.' and `Federation square'}
	\label{fig:sim}
\end{figure}

For example, the reference similarity between `Fed Sq.', as one of the place references for node \textit{b}, and a gazetteer entry `Federation Square' is calculated by first tokenizing both of the two strings into lists of tokens, and then measuring similarities for each pair of tokens. Finally, an average score all token-wise similarity is returned as the reference similarity, as shown in Figure~\ref{fig:sim}.

\subsubsection{Spatial similarity} \label{spatial}
All the obtained candidate entries within the derived ALR of a place are considered as satisfying all the stored spatial relationships to the neighboring anchor places with the same degree. Probabilistic uncertainties associated with each spatial relationship has not been discussed, i.e., any locations within an ALR are treated as of equal likelihood.  Spatial similarity is defined for measuring how well a gazetteer entry at a certain location satisfies the known spatial relationships, and features considered for measuring spatial similarity include orientation, distance, and topology. For example, if there are two squares as candidate entries that are obtained for the place reference `the large square' with exactly the same reference similarity, these two entries can only be further ranked considering their closeness to the anchor place `St Paul's Cathedral' given the spatial relationship \emph{near} in between.

Spatial similarity for a candidate gazetteer entry is computed as shown in Algorithm~\ref{ssim} by the average spatial similarity for all spatial relationships. Orientation and nearness similarities are calculated based on centroids of both the locatum and the relatum as points, while topological similarity is calculated based on polygons (if either the locatum or the relatum is geo-referenced not as a polygon, spatial similarity for this relationship will be ignored, i.e., continue the loop). 

\begin{algorithm}[ht]
	\caption{function calculateSpatialSimilarity()}
	\label{ssim} 
	\textbf{Input:} $p_{i}$, \emph{CandidateEntry}	    \Comment{All candidate entries for $p_{i}$ }	\\			
	\textbf{Output:} \emph{SpatialSimilarity}
	\begin{algorithmic}
		\State \emph{Similarities} $:= \emptyset$	
		\For {\emph{Relation}, \emph{AnchorPlace} \In getRelationshipToAnchorPlaces($p_{i}$)}	
		
			\\									\Comment{cardinal direction relationship}
			
			\If {\emph{Relation} \Is CardinalDirectionRelationship}
				\State \emph{Sim} = calculateOrientationSim(Relation)
			\\									\Comment{topological relationship}
			
			\ElsIf {\emph{Relation} \Is TopologicalRelationship}
				\State \emph{Sim} = calculateTopologicalSim(Relation)		
				\If {\emph{Sim} $=0$}			\Comment{filter out entries that do not satisfy}
					\State \Return $0$			\Comment{assign zero spatial similarity to the entry}
				\EndIf		
			\\									\Comment{qualitative distance relationship}
				
			\ElsIf {\emph{Relation} \Is QualitativeDistanceRelationship}
				\State \emph{Sim} = calculateNearnessSim(Relation)
				
			\\									\Comment{relative direction relationship}
			
			\ElsIf {\emph{Relation} \Is RelativeDirectionRelationship}
				\State \emph{NearnessSim} = calculateNearnessSim(Relation)
				\State \emph{OrientationSim} = calculateOrientationSim(Relation)
				\State \emph{Sim} = (NearnessSim + OrientationSim)/2
			
			\EndIf		
			\State \emph{Similarities} $\longleftarrow$ \emph{Sim}
		\EndFor
		\State \emph{SpatialSimilarity} $:=$ average(\emph{Similarities}) 
		\State \Return \emph{SpatialSimilarity}
	\end{algorithmic}
\end{algorithm}

Examples of spatial similarity calculation for three spatial relationships are shown in Figure~\ref{fig:ss} below. The shaded regions indicate search spaces. Nearness similarity is measured by (one minus) the distance between the centroid of the locatum and the relatum divided by the buffer distance, and must be between 0.0 and 1.0. Orientation similarity is measured by the angle between the displacement vector starting from the relatum to the locatum, and the direction vector pointing to the true direction (e.g., geographical north for \emph{north of}). Thus an orientation similarity must also be between 0.0 ($90^{\circ}$ for \emph{north, east, south, west}, and $45^{\circ}$ for composite directions such as \emph{northeast}) and 1.0 ($0^{\circ}$). Topology similarity is measured by whether the geometry of the locatum and the relatum satisfy the given topological relationship, and can be either 0.0 (not satisfy) or 1.0 (satisfy). The determination of topological relationship between two given polygons can be computed using existing algorithms and libraries.

\begin{figure}[ht]
	\centering
	\includegraphics[width=0.9\textwidth]{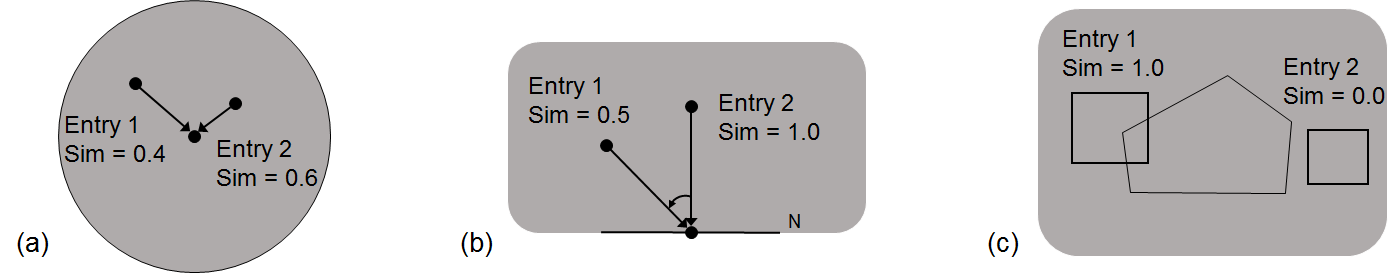}
	\caption{Spatial similarity calculation for \emph{near} (a), \emph{north of} (b), and \emph{overlap} (c)}
	\label{fig:ss}
\end{figure}

\subsubsection{Overall similarity scoring}
The overall similarity is calculated by function \emph{getOverallSimilarity} in Algorithm~\ref{BM} based on Eq.~\ref{eq:weight}. Different weights ($W_1$ and $W_2$) will be tested and evaluated in the implementation stage. Table~\ref{tab:sim} shows an example of calculating the overall similarities for three candidate entries for node \textit{b}. The highlighted cell indicates the candidate entry with the highest score, and this entry is used for geo-referencing node \emph{b}. Similarly, node \emph{e} (`Flinders Street Bake House') can also be geo-referenced.

\begin{equation}
\label{eq:weight}
OverallSim = W_1 \ast ReferenceSim + W_2 \ast SpatialSim
\end{equation}

\begin{table}[ht]
	\centering
	\begin{tabular}{cccc}
		\Xhline{4\arrayrulewidth}
		Place to be geo-referenced & Place reference & Candidate entry & Overall similarity \\
		\hline
		node \textit{b} & Fed Sq. & Ian Potter Centre & 0.17\\
		& & Federation Square & \cellcolor{gray!25}0.78\\
		& & Kirra Galleries & 0.20\\
		& the large square & Ian Potter Centre & 0.29\\
		& & Federation Square & 0.63\\
		& & Kirra Galleries & 0.30\\
		\Xhline{4\arrayrulewidth}
	\end{tabular}
	\caption{Example of best-matching for node \textit{b} based on computed overall similarities}
	\label{tab:sim}
\end{table}

\subsection{Geo-referencing non-gazetteered places} \label{non}
A non-gazetteered place is geo-referenced using its derived ALR. Thus, node \emph{f} from the sample graph can be geo-referenced as it is known that node \emph{f} is in front of node \emph{e} (`Flinders Street Bake House'), and node \emph{e} has already been geo-referenced by the last step through best-matching.

With such a representation, the location of the place can further be described using anchoring theory \cite{galton2005anchoring}. Thus, the place can be regarded as anchored to a location just by stating what is known with certainty and leaving the rest for further reasoning. Here the place can be described as \emph{anchored in} its derived ALR. 

For such a non-gazetteered place, if it has rich-connected spatial relationships to other anchor places, the derived ALR can be close to its actual location. Therefore, an alternative way is to use the centroid of an ALR to represent the approximate location of the place for geo-referencing purpose, and such a point-based representation can be more useful for applications such as geographic information retrieval. However, in this research we do not use the point-based representation as it is over-restricting. In comparison, using ALRs for geo-referencing non-gazetteered places preserves as much information as can be inferred without further generalization.

Currently there is no robust method to automatically distinguish gazetteered places without gazetteered place references (Situation 2) and non-gazetteered places (Situation 3) other than manual annotation. Therefore, in order to separate them, defining an overall similarity threshold is needed, i.e., classify all places geo-referenced by the best-matching process with overall similarities lower than the threshold as non-gazetteered places. Different threshold values will be tested and evaluated in case studies.

\section{Implementation and case studies}
The geo-referencing approach explained has been implemented in a system written in Python. Neo4j graph database\footnote{https://neo4j.com/} is used for storing place graphs as well as for querying. Some functions within the geo-referencing process such as geocoding by gazetteer, reverse-geocoding by region, Damerau-Levenshtein similarity calculation, WordNet-based semantic similarity measurement and geometry computation, are implemented using existing free Python packages. 

Three place graphs are tested. The first graph is constructed by Kim \emph{et al.} \cite{kim2015descriptions} using 44 descriptions (738 triplets in total) submitted by graduate students about the University of Melbourne campus. The other two place description datasets are harvested from web texts \cite{kim2015harvesting}, and are used to construct a place graph of Santa Fe, New Mexico (218 triplets) and a place graph of Melbourne, VIC (4173 triplets). Gazetteers used include OpenStreetMap, GeoNames and Google Maps.

Places from the three place graphs are manually annotated with labels \emph{anchor place} (places with at least one gazetteered place reference, Situation 1), \emph{gazetteered place} (places without gazetteered place references, but are still gazetteered, Situation 2), \emph{non-gazetteered place} (Situation 3) for evaluation purpose. Proportions of these places from the three input place graphs are shown in Table~\ref{table:category}. 

\begin{table}[ht]
	\centering
	\caption{Proportions of places from the three categories from the input place graphs}
	\label{table:category}
	\begin{tabularx}{0.95\textwidth}{l c c c}
		\Xhline{4\arrayrulewidth}
		Place graph & anchor place & gazetteered place & non-gazetteered place \\
		\hline
		Campus graph		& 28.6\% & 62.1\% & 9.3\% \\
		Melbourne graph	& 14.6\% & 71.9\% & 13.5\%  \\
		Santa Fe graph	& 19.7\% & 58.7\% & 21.6\%  \\
		\Xhline{4\arrayrulewidth}
	\end{tabularx}
\end{table}

\subsection{Geo-referencing anchor places}
\begin{figure}[ht]
	\centering
	\includegraphics[width=1\textwidth]{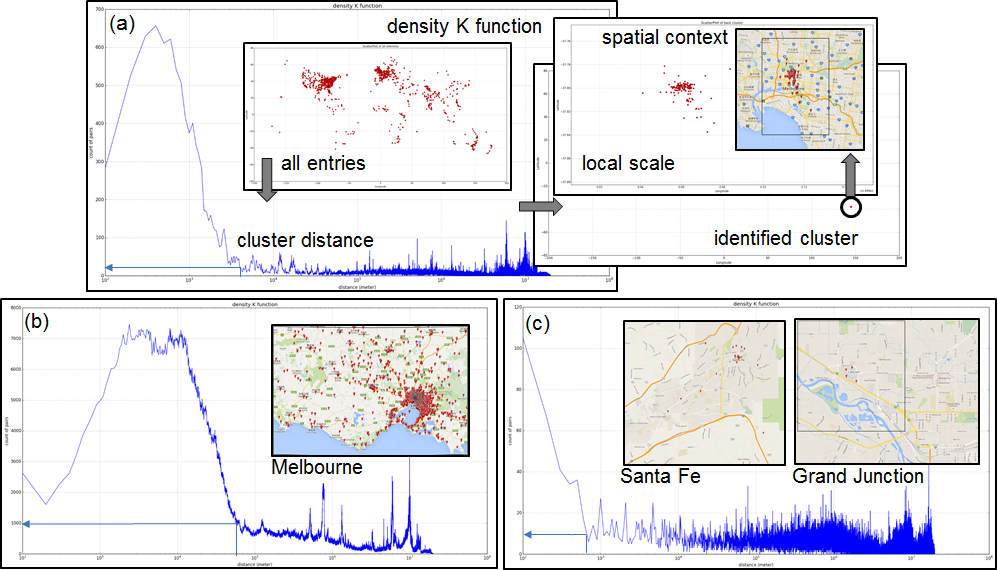}
	\caption{Clustering results and geo-referenced anchor places for the three place graphs}
	\label{fig:CR}
\end{figure}

Figure~\ref{fig:CR} shows the results of clustering as well as the disambiguated anchor places for the University of Melbourne campus graph (a), the Melbourne graph (b) and the Santa Fe graph (c). For each of the first two graphs, only one cluster is identified, and the cluster contains all the anchor places from the graph. For the Santa Fe graph, more than five clusters are identified, indicating multiple spatial foci. The top-ranking two clusters shown in Figure~\ref{fig:CR} (c) are within Santa Fe, New Mexico, US, and Grand Junction, CO, US respectively.

In order to measure how well these anchor places are geo-referenced, the standard evaluation metrics commonly used by toponym resolution studies are applied. Results are shown in Table~\ref{table:AP}. Precision is computed by the number of correctly geo-referenced anchor places divided by the total number of annotated anchor places. Recall values for anchor places are irrelevant since the same gazetteers are used for annotation as well as geo-referencing.

\begin{table}[ht]
	\centering
	\caption{Precisions for geo-referencing anchor places}
	\label{table:AP}
	\begin{tabularx}{0.85\textwidth}{l|c c c}
		\Xhline{4\arrayrulewidth}
		Place graph & Campus Graph & Melbourne Graph & Santa Fe Graph \\
		\hline
		Precision & 93.4\% & 87.5\% & 91.8\% \\
		\Xhline{4\arrayrulewidth}
	\end{tabularx}
\end{table}

\subsection{Geo-referencing through best-matching} 
\begin{figure}[ht]
	\centering
	\includegraphics[width=0.6\textwidth]{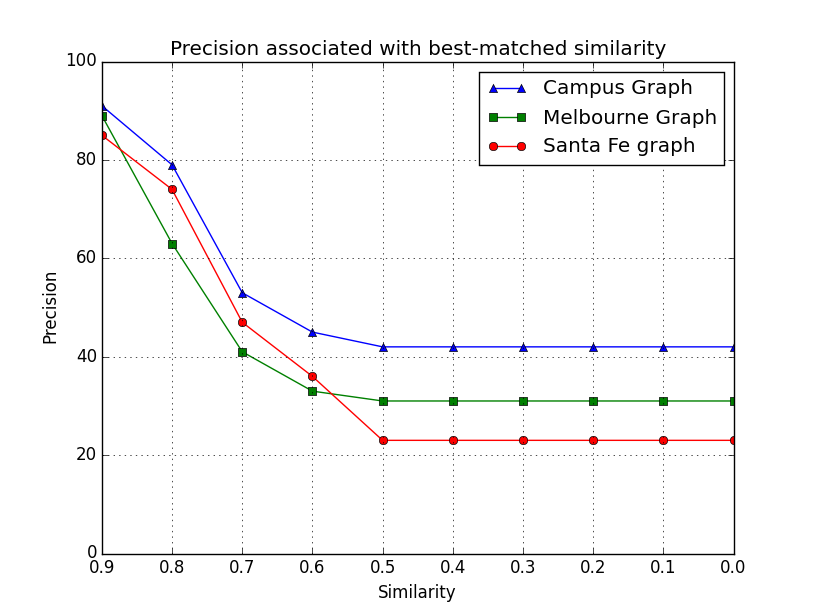}
	\caption{Precisions of best-matching associated with matched similarity}
	\label{fig:GA}
\end{figure}

For evaluating the best-matching process, all the places from the three place graphs that are annotated as \emph{gazetteered place} are considered. The result precisions for the three graphs are shown in Figure~\ref{fig:GA}. Each y-value represents the average precision for places matched with similarities greater than or equal to the x-value. For example, the y-value for the campus graph at 0.8 is 78.6\%, indicating that the overall precision for all best-matched places with similarities greater than or equals to 0.8 is 78.6\%.

Another evaluation metrics here is termed \emph{ALR precision}, and is defined as the number of places with their derived ALRs covering their corresponding gazetteer entries, divided by the total number of places annotated as \emph{gazetteered place}. A place can not be correctly geo-referenced if its derived ALR is not covering the location of the gazetteer entry. Overall precisions and ALR precisions for the three graphs are shown in Table~\ref{table:BM}.

\begin{table}[ht]
	\centering
	\caption{Precisions and ALR precisions for geo-referencing through best-matching}
	\label{table:BM}
	\begin{tabularx}{0.56\textwidth}{l c c c}
		\Xhline{4\arrayrulewidth}
		Place graph & Precision & ALR precision \\
		\hline
		Campus Graph 	& 42.7\% & 83.4\% \\
		Melbourne Graph & 31.1\% & 81.3\% \\	
		Santa Fe Graph 	& 23.1\% & 68.3\% \\
		\Xhline{4\arrayrulewidth}
	\end{tabularx}
\end{table}

As discussed before, different buffer distances for \emph{near} may lead to different geo-referencing results. Generally, a larger search space for \emph{near} is more likely to cover the true gazetteer entries of places to be matched, i.e., increase ALR precision, but at the same time increase the likelihood of getting false positives, i.e., decrease precision. Different parameters in Eq.~\ref{eq:near} are tested. However, no significant improvements in precisions and ALR precisions are observed. Reasons will be discussed later.

The best-matching algorithm is determined by both reference similarity as well as spatial similarity. Different weights are tested, and the overall precisions for the three place graphs are shown in Table~\ref{weight}. Previous results shown in Figure~\ref{fig:GA} and Table~\ref{table:BM} are based on the weights with the highest overall precision. Note that assigning different weights does not affect ALR precision, only precision.

\begin{table}[ht]
	\centering
	\caption{Precisions with different weights for best matching}
	\label{weight}
	\begin{tabularx}{0.7\textwidth}{c c c}
		\Xhline{4\arrayrulewidth}
		ReferenceSim weight & SpatialSim weight & Precision\\
		\hline
		1.0 & 0.0 	& 26.8\% \\	
		0.7 & 0.3 	& \cellcolor{gray!25}28.2\% \\	
		0.5 & 0.5 	& 14.8\% \\
		0.3 & 0.7 	& 8.1\% \\
		0.0 & 1.0 	& 2.2\% \\
		\Xhline{4\arrayrulewidth}
	\end{tabularx}
\end{table}

\subsection{Geo-referencing non-gazetteered places}
Place references referring to non-gazetteered places have no corresponding gazetteer entries, thus precision is not useful here and only ALR precision is considered. Results for the three input place graphs for non-gazetteered places are shown in Table~\ref{table:AN}.

\begin{table}[ht]
	\centering
	\caption{ALR precisions for non-gazetteered places}
	\label{table:AN}
	\begin{tabularx}{0.85\textwidth}{l|c c c}
		\Xhline{4\arrayrulewidth}
		Place graph & Campus Graph & Melbourne Graph & Santa Fe Graph \\
		\hline
		ALR Precision & 81.4\% & 77.5\% & 72.1\% \\
		\Xhline{4\arrayrulewidth}
	\end{tabularx}
\end{table}
 
Previous results are computed based on the manual annotated ground-truth labels. As already discussed in Section~\ref{non}, a threshold is necessary in order to automatically classify non-gazetteered places and gazetteered places. The evaluation metrics of classification is recall, and is defined as the proportion of places that are correctly classified for the two classes \emph{gazetteered place} and \emph{non-gazetteered place}. Applying different threshold values will affect the recall values for both two classes.

\begin{figure}[ht]
	\centering
	\includegraphics[width=0.6\textwidth]{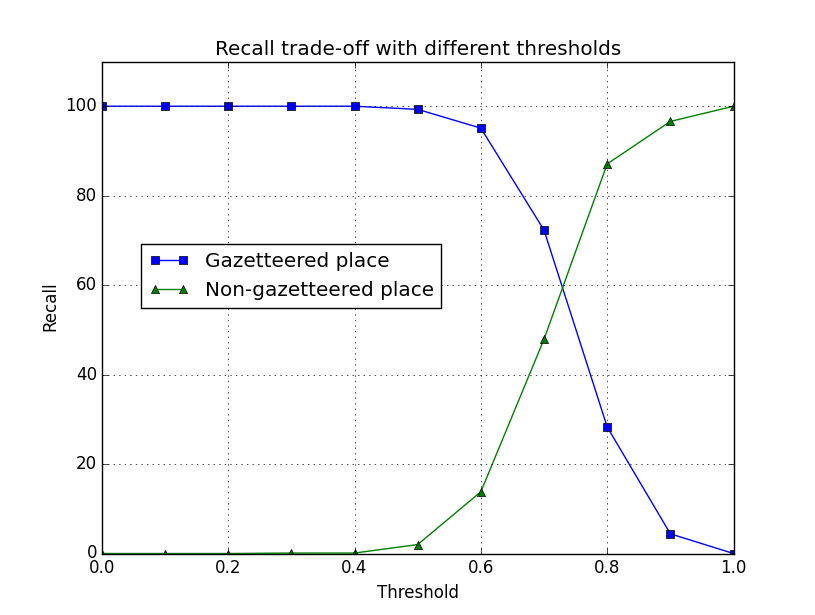}
	\caption{Recall trade-off between classes \emph{gazetteered place} and \emph{non-gazetteered place} with different threshold values}
	\label{fig:threshold}
\end{figure}

For example, if the threshold is set to 0.9, then most places here will be classified as \emph{non-gazetteered place}, and the recall value for \emph{gazetteered place} will be small consequently. Different thresholds are tested, and the overall recall values for the two classes for the three input graphs are shown in Table~\ref{fig:threshold}.

\section{Discussion}
The approach explained above is feasible for geo-referencing all places from everyday place descriptions, and its flexibility and applicability is demonstrated by case studies using place descriptions collected from different sources.

\subsection{Comparison and evaluation of the presented methodology}
The developed methodology is compared to the existing toponym resolution approaches and engines. However, the comparison is not straightforward since the objectives and tasks are quite different. Previous toponym resolution studies typically focus on gazetteered place references, referring to places of spatial granularities that are coarser (e.g., cities, countries, and geographic features). Such places are important for their objectives such as to determine the spatial foci of text documents for geographic information retrieval purposes, or to geo-reference places from historical document collections. For the task of this research, we are additionally interested in places of finer spatial granularities as well as places with non-gazetteered references. Such places are typically ignored in previous studies. Therefore, places from an input place graph are divided into three categories: \emph{anchor places}, \emph{gazetteered places}, and \emph{non-gazetteered places}, to allow a fair comparison.

The task of geo-referencing anchor places is comparable to existing toponym resolution approaches as anchor places are gazetteered places with gazetteered references. As shown in Table~\ref{table:AP}, the precisions for this task for the three tested place graphs are approximately 90\%. Existing toponym resolution engines, e.g., CLAVIN\footnote{http://clavin.berico.us/clavin-web/}, STEWARD\footnote{http://steward.umiacs.umd.edu}, or NewsStand\footnote{http://newsstand.umiacs.umd.edu}, have quite low recall for those anchor places due to the differences in the target corpora and gazetteers used, thus cannot be compare directly by testing our datasets. Therefore, we choose to compare our results to the most similar map-based toponym resolution approaches \cite{habib2012improving,buscaldi2008map,moncla2014geocoding} which have already been discussed and compared to our approach in Section~\ref{dis}, and the overall precisions for these approaches are from to 45\% to 94\%. Therefore, it is reasonable to say the approach developed for geo-referencing anchor places is as good as (if not better than) other existing toponym resolution approaches for the task of research in terms of precision.

As shown in Table~\ref{table:category}, the proportion of anchor places from the three datasets are approximately only 10\% to 30\%, while the remaining places do not have gazetteered references. For gazetteered places without gazetteered references, the overall precision and ALR precision of the proposed approach are shown in Table~\ref{table:BM}, indicating that the approach is feasible for geo-referencing these places. The ALR precision values are generally much higher than the precision values, which means that most places are covered by their derived ALRs, even if not all of them are successfully geo-referenced by their corresponding gazetteer entries by the best-matching process. Figure~\ref{fig:GA} shows that the geo-referencing precision is generally higher if the similarity scored by the best-matching process is higher, i.e., places matched with higher overall similarities are more likely to be correctly geo-referenced. For places that are matched with similarity values equal to or greater than 0.9, the overall precisions are approximately 90\%. Finally, ALR precision is also used for non-gazetteered places, and results are shown in Table~\ref{table:AN}. Precision is not used for non-gazetteered places since these places have no corresponding gazetteer entries. The result ALR precision values for non-gazetteered places are close to the ALR precisions for gazetteered places for the three input graphs (comparing Table~\ref{table:BM} and Table~\ref{table:AN}). This is because ALRs for places from both the two categories (\emph{gazetteered place} and \emph{non-gazetteered place}) are derived using the same method. Being able to geo-reference places without gazetteered references results in higher recall (and overall higher precision) compared to the results by existing toponym resolution approaches.

\subsection{Robustness, uncertainties, and parameter testing}
The approach for disambiguating and geo-referencing anchor places is based on a novel density-based clustering algorithm, as explained in Section~\ref{dis}. Three input place graphs that are of various sizes, different spatial scopes, and are collected from different sources, have been tested. The results reveal that the approach is feasible to geo-reference places from different input graphs. The algorithm is able to identify multiple spatial foci even if they are far away from each other. Taking the result for the Santa Fe place graph as an example (see Figure~\ref{fig:CR} (c)), multiple clusters are identified as corresponding to spatial foci, and anchor places within these clusters are disambiguated and geo-referenced successfully. The Santa Fe descriptions are harvested from websites and possibly contain places that are not in Santa Fe. 

Search spaces and spatial similarity are defined to determine the degree of satisfaction of candidate gazetteer entries given spatial relationships to some already geo-referenced anchor places. Search spaces are defined using existing models (e.g., half-plane models for cardinal direction relationships, and buffered regions for \emph{near}). Features including distance, orientation and topology are considered for measuring spatial similarity. For example, given a spatial relationship \emph{near} between a place to be matched and an anchor place, and two candidate gazetteer entries with the same name, the entry that is closer to the anchor place will be assigned with a higher spatial similarity. The major uncertainty comes from defining the buffer distance for \emph{near}, as discussed in Section~\ref{qualitative}, and the task remains an unsolved hard problem in relevant research fields such as geographic information retrieval. Here is a trade-off problem, i.e., larger buffer distances tend to result in higher ALR precision but at the same time increase the number of false positives. The buffer distance for \emph{near} in this research is defined considering both the size of the spatial context and the size of the referred relatum, as shown in Eq.~\ref{eq:near}, in order to capture at least some aspects of context. Different values of parameters in Eq.~\ref{eq:near} are used for tuning in the implementation stage, however no significant improvement in precision value is observed. A possible reason is that a large proportion of anchor places in the used gazetteers are geo-referenced as points, thus relatum sizes make no difference for these anchor places. Also, for two of the three tested graphs, only one spatial context is derived that contains all anchor places, thus the size of the spatial context makes no difference in such situation either.

The best-matching process, as described in Section~\ref{sec:bm}, considers reference as well as spatial similarity. Reference similarity is measured by a comprehensive algorithm considering token-wise string and semantic similarities and common abbreviations. Spatial similarity is already discussed in the previous paragraph. Different weights for the two components in order to compute the overall-similarity have been tested for tuning purposes, as shown in Table~\ref{weight}. The result shows that assigning reference similarity with weights around 0.7 gives the highest precision. Assigning reference similarity with a weight of 1.0 still gives quite high precision. In contrast, assigning spatial similarity with a weight of 1.0 results in nearly zero precision. A likely reason is that the obtained gazetteer entries for each place to be matched have already been filtered by spatial relationships, and reference similarity is more effective for further ranking these entries than spatial similarity.

The last step is to identify and separate out non-gazetteered places. Different thresholds have been tested for classification, and the result is shown in Table~\ref{fig:threshold}. The results reveal a trade-off pattern between the recall of gazetteered places and non-gazetteered places. Defining a threshold of around 0.7 results in the optimal situation, i.e., the overall highest recall values for both gazetteered and non-gazetteered places. 

\subsection{Failure analysis}
Three main reasons have been identified that cause incorrect geo-referencing of places. First, some derived ALRs are not covering the true locations of the corresponding places. This is most likely caused by inappropriate search spaces, e.g., too small buffer distance for \emph{near}. Incorrect ALRs affect both the best-matching precision (Table~\ref{table:BM}) for places annotated as \emph{gazetteered place} (not including anchor places), as well as the ALR precision (Tables~\ref{table:BM} and \ref{table:AN}). Second, gazetteered places with stored place references too different from the actual gazetteered names tend to have lower geo-referencing precision since they are difficult to be matched by the best-matching process. Third, some place references identified as gazetteered place names are actually not gazetteered. For instance, the place reference `Gate 10' referring to a non-gazetteered entrance of the University of Melbourne is identified as an anchor place because there is a gazetteer entry with the same name referring to another place. Such cases are rare, and are not considered by the proposed approach since they cannot be identified without further considering the sentence-level context of the original place description.

\section{Conclusion}
Natural language place descriptions occur in everyday verbal communication as a way of conveying spatial information about place. An important step of utilizing the contained knowledge from such place descriptions is to identify and geo-reference all places referred. Everyday place descriptions are flexible, vernacular, and often contain place references as synonyms or place categories, instead of officially stored place names in a gazetteer. Such place references are not known by a gazetteer, thus cannot be geo-referenced using current toponym resolution approaches which are typically based on gazetteered place name matching and disambiguation. In addition, place descriptions may also contain other places that have vague boundaries and can only be located by providing additional spatial relationships to other places, or places that are too fine-grained where environment features are no longer gazetteered (e.g., rooms). Even if some fine-grained places are gazetteered, they are usually more ambiguous and require different approaches to resolve other than some standard toponym resolution heuristics (e.g., based on population or jurisdictional containment relationship). Therefore, this research is motivated by developing a novel approach that could overcome these limitations and be able to geo-reference all places from place descriptions.

This research starts from a place graph which stores extracted place references and spatial relationships from any number of place descriptions, instead of focusing on a single document at a time. A place graph is constructed by spatial triplets extracted from place descriptions, and place references as synonyms are merged through a comprehensive similarity matching process. The complete methodology for place graph construction is provided in previous research \cite{kim2015descriptions,kim2016similarity}. The merged place references allow linking some non-gazetteered place references to gazetteered names, and the stored spatial relationships provide a qualitative reference system for describing places which can be used for constraining the locations of places.  

The proposed geo-referencing approach consists of three main stages. In the first stage, places in an input place graph that have at least some gazetteered place names are identified, disambiguated and geo-referenced. These places are then labelled as anchor places, and are used in the following steps to help geo-referencing the remaining places using spatial relationships. A novel density-based clustering algorithm is developed for this purpose, which is superior to comparable clustering approaches for the task of this research for several reasons (see Section~\ref{dis}). In the second step, for each of the remaining places, all its stored spatial relationships to the already geo-referenced anchor places are extracted and used to derive an approximate location region to constrain its likely location. Then, a comprehensive best-matching process is conducted based on comparing the stored place references and all gazetteer entries obtained within its derived approximate location region considering string, semantic, and spatial similarity. In the last stage, places that are non-gazetteered are identified and geo-referenced using their derived approximate location regions.

The developed approach has been tested with several datasets collected from different sources that are of various types (e.g., collected by survey or harvested from websites), sizes (constructed by potentially thousands of triplets), and spatial scopes (e.g., of different spatial granularities or with multiple spatial foci). The implementation was tested with over 5000 triplets in total, and the results show that the approach is feasible and applicable. In order to make a fair comparison to existing toponym resolution approache which typically only consider places with gazetteered place names, the evaluation is divided into three parts. For geo-referencing anchor places (place with gazetteered names), the novel approach has approximately 90\% precision for all the tested datasets (Table~\ref{table:AP}). However, the main contribution is that the developed approach is also able to geo-reference the remaining places (as shown in Table~\ref{table:category}, such places are approximately 70\% to 85\% in the input place graphs) that cannot be resolved using existing toponym resolution approaches, thus increase overall precision. The results show that approximately 20\% to 40\% of such places can be correctly geo-referenced (Table~\ref{table:BM}), and about 60\% to 80\% of these places are within their derived approximate location regions using spatial relationships (Table~\ref{table:BM} and \ref{table:AN}).

There are several adjustable parameters in the presented approach. Different parameter values have been tested in the implementation stage, and associated uncertainties and influences have already been discussed in the discussion section. For instance, defining the search space for \emph{near} is still an open question in relevant research fields (see Section~\ref{qualitative}) and heavily relies on contextual information and user perceptions. Such knowledge currently cannot be automatically extracted and modelled from place descriptions. In future work, refined search spaces for \emph{near} can be replaced to increase overall geo-referencing precision.

The applicability of the presented geo-referencing approach depends on both the richness of the input spatial relationships and place references. A richly-populated place graph tends to result in higher geo-referencing precision. As explained in Section~\ref{Place}, an input place graph may be derived from multiple place descriptions, e.g., a place graph of Melbourne can be constructed from hundreds of place descriptions about places in Melbourne. It is expected that as a place graph gets populated with more place descriptions and spatial relationships, the ALRs derived are expected to become more constrained and closer to the actual footprints thus increase the overall geo-referencing precision. Also, more non-gazetteered place references are likely to be merged with some gazetteered place names thus allow easier geo-referencing.

This research presents a feasible approach to geo-reference all referred places in everyday place descriptions. The outcome has potential benefits to various areas including geographic information retrieval which heavily relies on techniques that are able to automatically geo-reference places from text documents. Another application area is emergency services, which quickly fail when facing vernacular place descriptions with non-gazetteered place references and qualitative spatial relationships. The standard available geographic information systems (such as national address files) used in such situations are possibly not detailed enough for localization with regard to vernacular or granularity. Furthermore, the presented approach is able to enrich authoritative datasets, such as digital gazetteers and address databases, with people's local geographic knowledge. 

\bibliographystyle{ACM-Reference-Format-Journals}
\bibliography{bibfile}
\end{document}